\documentclass[sigconf]{acmart}
\usepackage{pifont}
\usepackage{multirow}
\usepackage{amsthm}
\usepackage{float}
\theoremstyle{definition}
\newtheorem{definition}{Definition}[section]

\AtBeginDocument{%
  \providecommand\BibTeX{{%
    \normalfont B\kern-0.5em{\scshape i\kern-0.25em b}\kern-0.8em\TeX}}}

\setcopyright{acmcopyright}
\copyrightyear{2024}
\acmYear{2024}
\acmDOI{XXXXXXX.XXXXXXX}

\acmConference[WSDM '24]{the 17th ACM International Conference on Web Search and Data Mining}{March 04--08,
  2024}{Mérida, México}
\begin{document}

\title{MolCPT: Molecule Continuous Prompt Tuning to Generalize Molecular Representation Learning}

\author{Cameron Diao}
\authornote{Both authors contributed equally to this research.}
\email{cameron.diao@rice.edu}
\author{Kaixiong Zhou}
\authornotemark[1]
\email{Kaixiong.Zhou@rice.edu}
\affiliation{%
  \institution{Rice University}
  \city{Houston}
  \state{Texas}
  \country{USA}
}

\author{Zirui Liu}
\email{zl105@rice.edu}
\affiliation{%
    \institution{Rice University}
    \city{Houston}
    \state{Texas}
    \country{USA}
}


\author{Xiao Huang}
\email{xiaohuang@comp.polyu.edu.hk}
\affiliation{%
  \institution{Hong Kong Polytechnic University}
  \city{Hong Kong}
  \country{China}
}

\author{Xia Hu}
\email{xia.hu@rice.edu}
\affiliation{%
  \institution{Rice University}
  \city{Houston}
  \state{Texas}
  \country{USA}
}

\renewcommand{\shortauthors}{Diao and Zhou, et al.}

\begin{abstract}
Molecular representation learning is crucial for accurately predicting molecular properties. Graph neural networks (GNNs) are effective methods for molecular representation learning, but training GNNs often needs large amounts of labeled molecular structures, which are expensive to obtain. Recently, the ``pre-train, fine-tune'' strategy has been adopted to improve the generalization capabilities of GNNs. It pre-trains GNNs in a self-supervised fashion and then fine-tuned to perform the downstream task with few labels. However, this strategy is not in line with molecular representation learning. Molecules are often characterized by motifs. Pre-training may not capture key structural information, and identifying suitable pre-training objectives often involves intensive engineering experimentations. To this end, we propose a novel strategy of ``pre-train, prompt, fine-tune'' for molecular representation learning. Our framework - molecule continuous prompt tuning (MolCPT), defines a prompting function that uses the pre-trained model to construct expressive prompts for each input molecule. The prompt effectively augments the molecular graph with informative motifs in the continuous representation space; this provides the pre-trained model with more structural patterns to aid downstream classification. In addition, we propose a differentiable answer search algorithm to map any format of pre-training output to the desired molecular property labels. Extensive experiments on several benchmark datasets show that MolCPT efficiently generalizes pre-trained GNNs for molecular property prediction, with or without a few fine-tuning steps. Our code can be found at: https://anonymous.4open.science/r/GraphCL-7105.
\end{abstract}

\ccsdesc[500]{Applied computing~Chemistry}
\ccsdesc[500]{Computing methodologies~Semi-supervised learning settings}
\ccsdesc[500]{Computing methodologies~Learning latent representations}
\ccsdesc[300]{Computing methodologies~Neural networks}

\keywords{Molecular property prediction, graph neural networks, self-supervised learning}



\maketitle

\section{Introduction}
Molecular property prediction is a fundamental task in many fields, such as quantum chemistry \citep{MPNN, SchNet}, drug discovery \citep{Stokes, DrugDiscoverySurvey}, and toxicity detection \citep{ToxicitySurvey}. Various graph neural networks (GNNs) have been proposed for molecular property prediction due to their effectiveness in modeling the chemical structure of molecular graphs \citep{MPNN, SchNet, MGC}. Specifically, they treat each molecule as a computation graph where the atoms of the molecule are the nodes and the bonds are the edges. The GNNs then learn informative low-dimensional embeddings for the nodes via message passing over the edges. Finally, the learned embeddings are pooled to form whole graph representations, which are used to predict the desired property.

Although GNNs have achieved remarkable success as molecular models, they are notorious for being data hungry, i.e., requiring sufficiently large amounts of labeled data. This is especially problematic for molecular representation learning, since labeled molecules typically occupy an extremely small portion of the entire chemical space \citep{MPPSurvey}. Furthermore, the labeling process is quite expensive, since labels can only be obtained using time-consuming wet lab experiments or quantum chemistry calculations. Thus, recent GNN studies have turned to self-supervised pretraining as a way of diminishing the need for labels \citep{Hu, GraphCL, Grover}. Self-supervised pre-training tasks (e.g., contrastive learning or masked prediction) can help GNNs learn general molecular structure even from unlabelled collections of molecules; the pre-training is expected to improve generalization to label-limited downstream tasks, with careful fine-tuning.

However, the ``pre-train, fine-tune'' strategy can fail to capture semantic structures instrumental for molecular analysis \citep{Sun}. Unlike social network graphs, molecules are often characterized by motifs, or frequently occurring subgraph patterns indicative of molecular properties. For example, Benzene ring is a functional motif of organic molecules that indicates aromaticity. Standard pre-training (e.g., comparing contrastive examples through random structure masking) is ill-suited for learning these informative motifs, which may lead to uninformative molecular representations. Even worse, \citet{Sun} report that self-supervised pretraining does not significantly improve performance without careful experimental setup. As a result, researchers must laboriously tailor the pre-training objective to the downstream task for desired results.

To avoid pre-training objective engineering, we investigate the strategy of ``pre-train, prompt, fine-tune'' for molecular representation learning. Prompting was first proposed in natural language processing (NLP) as a way of augmenting the input text with knowledge descriptions related to the downstream task \citep{PromptSurvey, PrefixTuning, OptiPrompt}. This can help the pre-trained model generalize effectively to a wide range of problem settings. To given an example, consider product review classification with raw input text ``Absolutely a cost-effective product.''. The prompting function reformulates the input text into a prompt by appending a task related description, such as ``Absolutely a cost effective product. Is the product good? \underline{[MASK]}''. This prompts the language model to produce the desired result by filling the masked word as in pre-training process. The masked word is then mapped to a positive or negative review evaluation to conduct the downstream classification applications. Motivated by the many successes of prompts in NLP, we aim to define a \textit{motif} prompting function that aids pre-trained GNNs in molecular analysis.

Unfortunately, it is non-trivial to design the motif prompting function due to the following two challenges: (i) It is unclear how to construct the prompts, which involves augmenting the molecular graph with task-related motifs. Unlike sequential text, graphs are comprised of unordered nodes and their physical connections. No canonical node ordering makes it difficult to combine the molecule and motifs in a well-defined, meaningful way. (ii) Second, it is challenging to design the answer search strategy that maps the pre-training output to the downstream molecular properties. While one can search labels from filled words based on the semantic similarity of words, it is infeasible to map the pre-trained graph topology knowledge to downstream molecular labels correspondingly.


To tackle the above challenges, we propose molecule continuous prompt tuning (MolCPT) for enhancing downstream molecular analyses. MolCPT adopts the ``pre-train, prompt, fine-tune'' strategy by first pre-training the backbone GNNs using self-supervised learning. Then, it employs a novel motif prompting function to augment the molecular graph with motifs in the continuous representation space. This forms a motif prompt that helps the pre-trained GNNs easily identify the underlying molecular properties. In addition, we design a differentiable answer search algorithm to map any format of pre-training output to the molecular property labels. Through MolCPT, we make three key contributions to the field of molecular property prediction:

\begin{itemize}
    \item To flexibly prompt molecular representations (\textbf{challenge 1}), our motif prompting function combines the molecular and motif representations in continuous representation space. This removes the requirement that the motifs should be structurally connected to the molecule in discrete chemical space, which requires specific domain knowledge. Furthermore, we train our prompting function to denoise the motif prompts and encode them with task-relevant information.
    \item We propose a differentiable answer search scheme for the molecular property prediction (\textbf{challenge 2}), which accompanies contrastive and node masking pre-training tasks. The idea is to construct a set of trainable answer vectors to represent the downstream labels, and perform classification by estimating pair-wise similarities between the pre-training output and the answer vectors. Our scheme enables the diverse GNNs pre-trained with different tasks to be reused for the molecular property prediction.

    \item We evaluate MolCPT on a series of molecular benchmark datasets. Our experimental results demonstrate that motif prompting efficiently generalizes pre-trained GNNs to a number of downstream tasks. In particular, MolCPT beats pre-training baseline scores on 5 out of 6 of the datasets.
    
\end{itemize}

\section{Preliminaries}

\subsection{Molecular Property Prediction}
A molecule is represented as a topological graph $G=(\mathcal{V,E})$, where $\mathcal{V}$ and $\mathcal{E}$ denote the atoms (nodes) and bonds (edges) of the molecule, respectively. Considering node $v\in \mathcal{V}$, we use $x_v\in \mathbb{R}^{d}$ to denote the node's initial features, and $\mathcal{N}_v$ to denote the set of its direct neighbors. Let $\mathcal{T} = \{(G, y), \cdots\}$ denote the training set of molecule-and-label pairs. The molecular property prediction task aims to learn informative molecular representations and map each one to their correct label. The molecular representation scheme should generalize over the extensive chemical space even with a small training set of labeled samples. 

\subsection{Graph Neural Networks}
GNNs learn the molecule's topological structure and atom features via hidden node embeddings, which are read out to generate the molecular representation. Specifically, following the message passing paradigm, GNNs learn node embeddings by recursively aggregating messages across direct neighborhoods (optionally including self-messages). At the $k$-th layer, the embedding learning function with input node $v$ is formulated as:
\begin{equation}
\label{eq: message_passing}
x^{(k)}_v  = \mathrm{AGGRE}\left(\left\{ x^{(k-1)}_{v'}, v'\in \mathcal{N}_v\cup v\right\}, \theta^{(k)} \right).
\end{equation}
$x^{(k)}_v\in \mathbb{R}^{d}$ is node $v$'s embedding vector at the $k$-th layer; $x^{(0)}_v = x_v$ is $v$'s embedding vector at the initial layer; $\theta^{(k)}\in \mathbb{R}^{d\times d}$ is a trainable weight matrix for encoding atom features; and $\mathrm{AGGRE}$ denotes the aggregation and combination function for messages, in this case node embeddings (e.g., through sum, mean, or max pooling). Suppose the number of graph convolutional layers is $K$. Then we use $x^{(K)}_v \triangleq f_{\theta}(G, v)$ to denote the final node embedding learned by the $K$-th layer, where $\theta=\{\theta^{(1)}, \cdots, \theta^{(K)}\}$ is the complete set of trainable parameters. 

To obtain the molecular representation used for property prediction, a readout function (e.g., sum, mean, or max pooling) is used to combine all of the node embeddings as: $h_G  \triangleq f_{\theta}(G) = \mathrm{READOUT}(\{x^{(K)}_v, v\in \mathcal{V}\})$.

\subsection{Pre-train and Fine-tune}
\label{sec: pre-train}
Pre-training methods are adopted to learn robust and transferable features of molecular graphs. The majority of these methods belong to one of two categories: predictive learning or contrastive learning. In predictive learning \citep{Hu, Grover}, we randomly mask part of the molecule, then pre-train the GNNs to recover that part. In contrastive learning \cite{SimCLR, GraphCL, MolCLR}, we maximize the mutual information between the original molecule and its masked counterpart.

The pre-trained model $f_\theta$ serves to initialize finetuning on the downstream task. Mathematically, the model $f_\theta$ is connected with a new classifier $p_\varphi$, often a multi-layer perceptron (MLP) with parameters $\varphi$. They are fine-tuned together as:
\begin{equation}
    \label{eq: tradition_tune}
    \begin{array}{cl}
    \min_{\theta, \varphi} & \sum \limits_{(G, y)\in \mathcal{T}}  \mathcal{L}(p_\varphi(f_{\theta}(G)); y), \\
    \mathrm{s.t.} & \theta^{\mathrm{init}} = \theta^{\mathrm{pre}}. 
    \end{array}
\end{equation}
The constraint means that the GNNs are initialized using the pre-trained model parameters. $\mathcal{L}$ is the classification loss function, such as cross-entroy loss. 

\begin{figure}[t]
    \centering
    \includegraphics[width=0.9\linewidth]{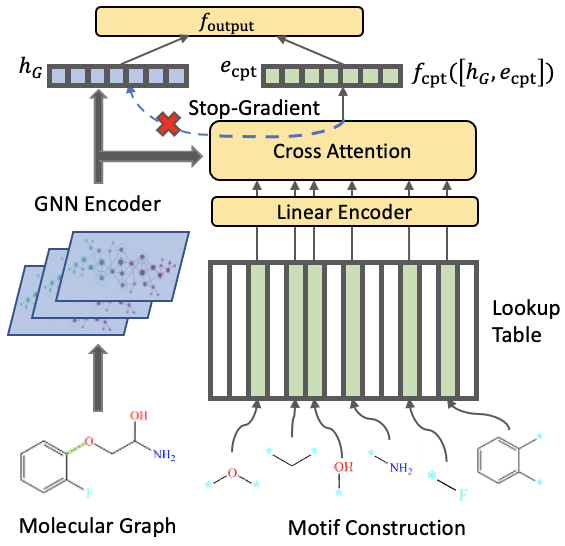}
    \caption{Overview of MolCPT. Molecular graph $G$ passes through the GNN encoder to obtain $h_{G}$. Concurrently, $G$ is fragmented into several motifs. MolCPT looks up the motifs' embeddings, then passes them through an attention network to obtain $e_{\mathrm{cpt}}$. $h_{G}$ and $e_{\mathrm{cpt}}$ are combined to produce a motif prompt for downstream molecular analysis.
    }
    \label{fig:molcpt_diagram}
    
\end{figure}

\section{Molecule Continuous Prompt Tuning}
\label{sec: method_summary}
In the molecular setting, pre-training methods often fail to improve downstream performance without extensive hyperparameter tuning and specific experiment settings. In particular, \citet{Sun} find that self-supervised pre-training only achieves marginal improvement or worse performance compared to no pre-training. This may be because self-supervised pre-training mainly relies on random context masking, and does not consider important biophysical properties expressed by, e.g., functional motifs \citep{FunctionalGroupImp}. It is worth studying how to use functional motifs to guide the pre-trained model in learning transferable, chemically significant features.

In this work, we propose using motifs to ``pre-train, prompt, and fine-tune''. In NLP, the prompting strategy adapts a single pre-trained model to diverse downstream tasks. This frees one from the labor of designing a specific pretraining objective for each task, and allows one to recycle previous models from the literature. Our proposed MolCPT leverages motifs to prompt input molecules, and adapt the pre-trained GNNs to each property prediction task.

\subsection{Pre-train, Prompt, Fine-Tune}
Before diving into the technical details of MolCPT, we first mathematically define the motif prompting function, then describe the entire pipeline of ``pre-train, prompt, fine-tune". 

\begin{definition}[Graph Motif]
    Consider molecular graph $G = (\mathcal{V}, \mathcal{E})$. We define its set of motifs as:
        
    \begin{equation}
        \mathcal{M}_G \triangleq \{M^{(1)}, \cdots, M^{(n)}\},
    \end{equation}
    \noindent where $M^{(j)}$ denotes the $j$-th motif, and $n$ denotes the total number of motifs belonging to molecule $G$. Since motifs are defined as subgraphs of $G$, each motif $M^{(j)} = (\mathcal{V}^{(j)}, \mathcal{E}^{(j)})$ where $\mathcal{V}^{(j)} \subset \mathcal{V}$ and $\mathcal{E}^{(j)} \subset \mathcal{E}$. 
\end{definition}
\label{graph_motifs}

\begin{definition}[Motif Prompting Function]
    A motif prompting function $f_{\mathrm{prompt}}$ is used to reformulate the input molecule by appending a series of associated motifs:
    \begin{equation}
            G' = f_{\mathrm{prompt}}(G, \mathcal{M}_G),
            \label{eq:motif_prompt}
    \end{equation}
    where $G'$ is a motif prompt, i.e., a hypergraph generated by highlighting the motifs that inform key molecular properties. 
\end{definition}

Several possible solutions can be adopted to define $f_{\mathrm{prompt}}$. For example, one can simply concatenate the inputs as disjoint graphs, or define rules to specify the connections between the motifs and their molecules. Our MolCPT solution (illustrated in Figure \ref{fig:molcpt_diagram}) opts for a different approach, and can be described by the following four steps:
\begin{itemize}
    \item \textbf{Model pre-training.} As introduced in Section~\ref{sec: pre-train}, we use a self-supervised pre-training objective to train the backbone GNN. We do not extensively design and pre-train task-specific models for each prediction task.
    \item \textbf{Prompt addition.} Following Eq.~\eqref{eq:motif_prompt}, we reformulate each standalone molecule $G$ into a motif prompt $G'$ by incorporating indicative motifs.
    \item \textbf{Answer search.} We construct differentiable answer vectors to represent labels in downstream classification problem. Based on the pre-training output in the latent space, we search the appropriate label by comparing it with the answer vectors one by one.   
    \item \textbf{Prompt fine-tuning.} We use motif prompt $G'$ to replace the original graph $G$ in the fine-tuning objective. Model parameters can be fine-tuned as in eq.~\eqref{eq: tradition_tune}, or by using the pre-training loss; we describe the procedure for ``fine-tuning like you pre-train'' in Section \ref{answer_search}.
\end{itemize}

These steps were designed to meet the challenges in defining an expressive prompting function for molecular property prediction (challenges outlined in the introduction). Specifically, our design comprises of three functional modules: (i) motif corpus generation to extract meaningful molecular subgraphs, (ii) continuous prompting function to reformulate the standalone molecule, and (iii) motif constrained learning to fine-tune the model.

\subsection{Motif Corpus Generation}

Real-world molecular datasets contain an enormous number of possible motifs, making it difficult to leverage these significant molecular structures. MolCPT preprocesses the molecular dataset to construct a very limited vocabulary of motifs, using chemical domain knowledge. The vocabulary acts as a prompting corpus for guiding molecular representation learning.

The motif vocabulary should meet two conditions: (i) The motifs must contain semantic structure information, e.g. by extracting meaningful functional groups from the molecular dataset. (ii) The motifs must occur frequently enough for stable training. We designed MolCPT to meet these conditions by using the graph fragmentation method provided by \citet{MGSSL}. Specifically, MolCPT adapts the Breaking of Retrosynthetically Interesting Chemical Substructures (BRICS) algorithm \citep{BRICS} to form the motif vocabulary. By further filtering out low-frequency motifs, we mathematically define the vocabulary as follows:

\begin{definition}[Motif Vocabulary]
    Considering dataset $\mathcal{T} = \{(G, y), \\ \cdots\}$, let $\mathcal{M}_G$ denote the set of fragmented motifs from graph $G$. A motif vocabulary is the union of frequent motifs:
        \begin{equation}
            \mathcal{M}_{\mathrm{mol}} \triangleq \left(\bigcup \limits_{G\in \mathcal{T}} \{M^{(j)}\in \mathcal{M}_G: |M^{(j)}| \ge t\}\right) \cup \{M^{(0)}\},
        \end{equation}
    \label{def:vocab}
    where $t$ thresholds the minimum motif frequency, and $\lvert M^{(j)} \rvert$ is the number of molecules in dataset $\mathcal{T}$ containing motif $M^{(j)}$. $M^{(0)}$ is an empty motif containing no nodes or edges: $M^{(0)}$ is assigned to molecules in $\mathcal{T}$ with no frequent motifs associated to them.
\end{definition}

\subsection{Continuous Prompting Function}

Prompting in the molecular domain is not straightforward as prompting in NLP, where one constructs prompts by directly concatenating discrete words. This is because naively connecting motifs and molecules as discrete graphs corrupts the molecular structure and leads to noisy molecular representations. Some works apply deterministic rules to explicitly connect the graphs, but these works rely heavily on domain knowledge, or worsen training efficiency due to the large number of candidate links \citep{MolDQN, GCPN, MolecularRNN}.

Instead of optimizing over discrete graph structure, we propose a continuous prompting function that connects motifs and molecules in the continuous representation space. The function is defined as: 

\begin{definition}[Continuous Prompting Function]
    A continuous prompting function $f_{\mathrm{cpt}}$ concatenates the embeddings of molecule $G$ and its frequent motifs, generating the prompt embedding as:
    \begin{equation}
        h'_G = f_{\mathrm{cpt}}(h_G, \{e^{(j)}, \mathrm{for}\ M^{(j)}\in \mathcal{M}_G\cap\mathcal{M}_{\mathrm{mol}}  \}).
        \label{eq:con_motif_prompt}
    \end{equation}
    \label{def: con_motif_prompt}
    $h_G$ and $h'_G$ are the original and final prompt embeddings of molecular graph $G$, respectively. $e^{(j)}\in \mathbb{R}^{d}$ is the embedding of motif $M^{(j)}$, which can be one-hot. We consider only the frequently appearing motifs, defined by the intersection set of $\mathcal{M}_G\cap\mathcal{M}_{\mathrm{mol}}$. To obtain the expressive prompt embedding $h'_G$, we store the trainable motif embeddings in a lookup table, and parameterize the continuous prompting function with an attention network.
\end{definition}

\subsubsection{Motif Embedding Table.}
\label{embedding_table}
We adopt a lookup table to store the motif embeddings, which are fine-tuned to encode semantic structures. Mathematically, we define the table as $\mathbf{E}_{\mathrm{mol}}\in\mathbb{R}^{\lvert \mathcal{M}_{\mathrm{mol}}\rvert \times d}$, where $\lvert \mathcal{M}_{\mathrm{mol}}|$ denotes the cardinality of the motif vocabulary.

The motif embeddings must be properly initialized to encode semantic structures. One naive strategy is to randomly initialize the embeddings and rely on training to learn these structures. However, random initialization may inaccurately distinguish between motifs at the initial stage and converge to poor generalization areas \citep{PrefixTuning}. Instead, we propose to initialize the motif embeddings by inferring them using the pre-trained GNNs. Recall from Definition \ref{graph_motifs} that each motif $M^{(j)}$ has a subgraph structure $(\mathcal{V}^{(j)}, \mathcal{E}^{(j)})$. The pre-trained GNNs take this subgraph as input to infer the motif embedding $e^{(j)}$ as: $e^{(j)} \triangleq f_\theta(M^{(j)})$. We initialize the empty motif embedding $e^{(0)}$ as all-zeros vector represents semantic absence. 

\subsubsection{Molecule-Motif Cross Attention}
Diverse motifs influence molecular properties differently, so the prompting function $f_{\mathrm{cpt}}$ should weight motifs based on how they influence these properties. However, in its current formulation, $f_{\mathrm{cpt}}$ is poorly suited to inferring these interrelations between motifs and molecules, since it separates them out as distinct graphs.

To address this problem, we use a multi-head attention module \citep{VaswaniTransformer} to weight the motifs and generate a single abstract motif representation for each molecule. Mathematically, let $\mathbf{E}_G$ denote the motif embedding matrix formed from the motifs in $\mathcal{M}_G\cap\mathcal{M}_{\mathrm{mol}}$, by stacking the embeddings row-wise. Then the abstract motif representation $e_{\mathrm{cpt}}\in \mathbb{R}^{d}$ used to prompt molecule $G$ is given by:

\begin{equation}
e_{\mathrm{cpt}} = \mathrm{softmax} \left(\frac{ (\mathbf{W}_q h_G)^\top (\mathbf{E}_G\mathbf{W}_k)}{\sqrt{d}}\right) \mathbf{E}_G\mathbf{W}_v, 
\label{eqn:cross_attn}
\end{equation}
where $\mathbf{W}_q$, $\mathbf{W}_k$ and $\mathbf{W}_v\in \mathbb{R}^{d\times d}$ are the query, key, and value transformation matrices, respectively. For clarity, we omit feedforward layers (along with residual connections and layer normalization) from Eq. \eqref{eqn:cross_attn} and only consider the single-head case. Eq. \eqref{eqn:cross_attn} treat the molecular representation $h_G$ as a query that attends to the associated motifs, producing the abstract motif representation $e_{\mathrm{cpt}}$. From $e_{\mathrm{cpt}}$, the prompting function $f_{\mathrm{cpt}}$ in Eq. ~\eqref{eq:con_motif_prompt} instantiates the final prompt embedding as: $h'_G = h_G + e_{\mathrm{cpt}} \in \mathbb{R}^{d}$. $h'_G$ is used as input to classifier $p_\varphi$ for property prediction.

As a final note, we freeze the backwards gradients on molecular representation $h_G$ when computing Eq. \eqref{eqn:cross_attn}. This decouples the gradient computations of the molecular and motif representations, with two advantages: (i) MolCPT can flexibly learn the informative motif prompts without restricting to or affecting knowledge held in the pre-tained GNN. (ii) The decoupling allows one to infer directly using the pre-trained GNN weights, and only fine-tune weights parameterizing the prompting function and classifier. This makes motif prompting an efficient plug-in technique for extending a pre-trained model to multiple downstream tasks.


\subsection{Differentiable Answer Search}
\label{answer_search}
Following Eq. \eqref{eq: tradition_tune}, standard fine-tuning requires replacing the prediction layer, since the pre-training objective is often different from the downstream problem. For example, one applies contrastive learning or node masking during pre-training but focuses on molecular property prediction, as studied in this paper. Prior work demonstrate that this objective shift prevents the pre-trained model from effectively adapting to the downstream problem \citep{FLYP, GPPT}. Even worse, a large number of labels is required to train the new prediction layer for desired generalization. 


To avoid objective shift, answer search has been proposed under the framework of  ``pre-train, prompt, fine-tune”, to perform downstream tasks in the same manner as pre-training. Consider a language model pre-trained with masked word prediction. Under answer search, all downstream problems are unified as word filling tasks (e.g., the product review prompt ``Absolutely a cost effective product. Is it good?  \underline{[MASK]}”). The language model fills in the masked word (e.g., ``Good”) and searches for the desired label (e.g., Positive) to conduct classification. However, it is difficult to design an answer searching scheme for molecular property prediction due to two main challenges: (i) While it is easy to search for answers based on the semantic similarity of words, it is infeasible to map the pre-trained graph topology knowledge to downstream molecular labels. (ii) The graph pre-training tasks are diverse in literature, so it is impossible to apply a fixed answer searching scheme to all the pre-trained models. 


Instead, we propose a differentiable answer search to map pre-training outputs to downstream labels in the latent embedding space. The idea is to construct a set of answer vectors to represent the downstream labels, and perform classification by estimating pair-wise similarities between the pre-training output and the answer vectors. Mathematically, let $\mathbf{Y}_{\mathrm{ans}} \in \mathbb{R}^{m\times d}$ be the set of answer vectors, which are properly initialized to represent labels as described later. Each answer vector is associated with one of $k$ labels.  In this work, we associate multiple answer vectors to each label to enhance the label representation capability. 
Let $\mathcal{Y}^{(i)}$ be the set of answer vectors associated with label $i$, for $i \in \{1, \ldots k\}$, and let $y^{(i)}$ denote the practical answer vector to measure similarity with pre-training output. Particularly, while $y^{(i)}$ is chosen uniformly from the set of answer vectors $\mathcal{Y}^{(i)}$ during training, it is obtained from mean average during inference as $y^{(i)} = \mathrm{MEAN}(\{y \in \mathcal{Y}^{(i)}\})$.
Then, given the continuous prompt embedding $h’_G$, it is forwarded into the pre-training output layer to obtain $y’_G = f_{\mathrm{output}}(h’_G)$. We issue label by finding the maximum similarity based on inner product between practical answer vector and pre-training output:
\begin{equation}
    \mathrm{argmax}_{i\in\{1, \ldots k\}} (y^{(i)\top} y'_{G}). 
\end{equation}




The graph pre-training is often based on link prediction, contrastive learning, or node masking, where the former two tasks are mathematically the same by measuring the similarity of a pair of views. We consider the contrastive learning and node masking settings to describe how to conduct answer search in detail.


\begin{figure}[t]
    \centering
    \includegraphics[width=0.9\linewidth]{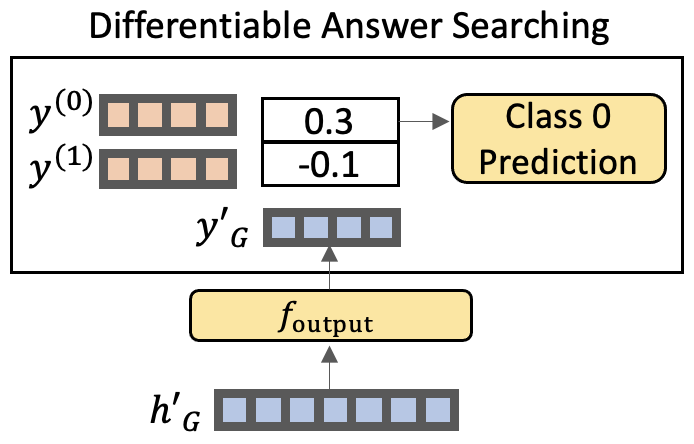}
    \caption{Differentiable answer searching for binary classification. Continuous prompt embedding $h'_G$ gets passed through the output layer $f_{\mathrm{output}}$ to obtain output $y'_G$. We compute pair-wise similarities between $y'_G$ and answer vectors $y^{(0)}$ and $y^{(1)}$ to obtain class probabilities. Finally, we predict the class with the highest probability, in this case class 0.}
    \label{fig:molcpt_diagram}
    
\end{figure}

\subsubsection{Contrastive Learning}
In the biochemical graph constrastive learning, GNNs are pre-trained to maximize agreement between augmented views of the same graph in a shared latent space. The graph embeddings $h_G$ are projected into this latent space using an MLP layer $f_{\mathrm{output}}$. During fine-tuning, given the prompt embedding $h'_G$ we reuse the MLP layer $f_{\mathrm{output}}$ to obtain $d$-dimensional output $y'_G = f_{\mathrm{output}}(h'_G)$. Following the same objective in pre-training, the model is trained to maximize agreement between $y'_G$ and answer vector $y^{(i)}$ with the same label $i$ as $G$. Before training, we initialize $y^{(i)}$ as $y^{(i)} = \mathrm{MEAN} (\{y'_G: G \in \mathcal{T}_{i} \} )$, where $\mathcal{T}_{i}$ is the subset of molecular graphs in dataset $\mathcal{T}$ with label $i$. This way, the answer vector agrees most with graph embeddings labeled as $i$.

\subsubsection{Node Masking}
\citet{Hu} propose the node masking based pre-training task, which trains GNNs to predict categorical attributes of masked nodes. Suppose the task is to predict the atom type of a node, out of $c$ possible types. To do so, we use linear layer $f_{\mathrm{output}}$ to transform the node embedding into a $c$-dimensional logit vector for prediction. During fine-tuning, we reuse $f_{\mathrm{output}}$ to transform prompt embedding $h'_G$ into a $c$-dimensional logit vector $y'_G$. $y'_G$ is compared pair-wisely with answer vectors $y^{(i)}$ for $i \in \{1, \ldots k\}$ (Note: Herein we initialize the dimension size of $y^{(i)}$ as $c$ to facilitate the comparison with the logit vector). As a consequence, we obtain a $k$-dimensional similarity score vector represents the probability distribution over the label space in the downstream molecular property classification.
The model normalizes the $k$-dimensional similarity score vector using softmax function, and then uses it to compute cross entropy loss. 

\subsubsection{Fine-tuning with Prompt}

While the proposed motif prompt and differentiable answer search can remove the objective shift between pre-training and fine-tuning, it may underperform when those prompt related parameters are unsuited to learning the domain-specific features. Therefore, compared with traditional fine-tuning in Eq.~\eqref{eq: tradition_tune}, we have the following optimization problem:

\begin{equation}
    \begin{array}{cl}
    \min \limits_{\theta,\mathbf{E}_{\mathrm{G}}, \mathbf{W}_{q, k, v}, \mathbf{Y}_{\mathrm{ans}}} & \sum \limits_{(G, y) \in \mathcal{T}} \mathcal{L}_{\mathrm{pre}}(f_{\mathrm{output}}(h'_{G}); y) \\
    \mathrm{s.t.} & \theta^{\mathrm{init}} = \theta^{\mathrm{pre}}.
    \end{array}
\end{equation}

\noindent where $\mathcal{L}_{\mathrm{pre}}$ is the pre-training loss, which is reused based on the differentiable answer search. The pre-trained GNNs $f_{\theta}$ can be frozen during fine-tuning for easy deployment of the GNNs to multiple downstream tasks. Then, MolCPT is simplified to tuning $\mathbf{E}_{\mathrm{ans}}, \mathbf{E}_{\mathrm{mol}}, \mathbf{W}_{q, k, v}$.

\section{Related Work}


\noindent\textbf{Pre-training GNNs.} GNN pre-training aims to capture structural patterns in the input graph distribution, often in a self-supervised manner. A growing number of pre-training strategies have demonstrated great success at transfer learning. For example, Deep Graph Infomax \citep{DGI} pre-trains GNNs to maximize mutual information between subgraph representations and high-level graph summaries. \citet{Hu} pre-trains on the level of individual nodes as well as entire graphs. GraphCL \citep{GraphCL} uses contrastive learning with domain-specific graph augmentations, significantly improving transfer learning on molecular tasks. Certain pre-training strategies are designed specifically for learning molecular features. MGSSL~\cite{MGSSL} introduces a motif generation task for capturing substructures in molecular graphs. MolCLR~\cite{MolCLR} uses a contrastive learning framework that encodes general molecular structure. Unfortunately, \citet{Sun} found that the most popular pre-training strategies do not significant improve transfer learning performance on molecular tasks. This work aims to guide pre-trained GNNs in learning more informative molecular features.

\noindent\textbf{Prompt in Natural Language Processing.} Recent works in NLP have adopted a new strategy for applying pre-trained language models to downstream tasks, called ``pre-train, prompt, fine-tune" \citep{PromptSurvey}. This strategy does not use a task-specific pre-training objective, but instead defines a task-specific prompting function that aligns the original pre-training task with the help of textual prompts. Prompts can be engineered for better downstream performance in two different ways. Manual Template Engineering creates prompt templates from human intuition \citep{PET}. Automated Template Learning learns prompt templates as part of the finetuning procedure, and can learn over discrete \citep{AutoPrompt, LM-BFF} or continuous prompts \citep{MixturePrompts}. 

This work takes inspiration from two prompt-related works, GPPT \citep{GPPT} and FLYP \citep{FLYP}. Like GPPT we recycle the pre-trained output layer for answer searching on the downstream task, and like FLYP we perform fine-tuning using the pre-training loss. However, GPPT cannot be straightforwardly applied to graph classification; we introduced prompt ensembling and a zero-shot prediction procedure to beat standard fine-tuning performance. Similarly, FLYP was designed for text-image classification, and cannot be applied to the molecular domain where manually designed prompts are practically infeasible.

\begin{table*}[t]
    \setlength{\tabcolsep}{6pt}
        \centering
        \caption{Results evaluating models on six MoleculeNet datasets, averaged across 5 runs and measured in ROC-AUC (\%). MolCPT beats pre-trained baselines on 5 out of 6 classification tasks. Average (Avg.) raking denotes the overall ranking of an algorithm, where the smaller ranking is the better. } 
        \begin{tabular}{ll|ccccccccc}
        \toprule
        \multirow{2}*{Frameworks} & \multirow{2}*{Models} & \multicolumn{6}{c}{Datasets} & \\
         &  & \textbf{BBBP} & \textbf{BACE} & \textbf{ClinTox} & \textbf{Tox21} & \textbf{HIV} & \textbf{SIDER} & Avg. ranking \\
        \midrule
        \multirow{4}*{Supervised} & GCN~\cite{GCN} & 64.9\textcolor{gray}{$\pm$3.0}&73.6\textcolor{gray}{$\pm$3.0}&65.8\textcolor{gray}{$\pm$4.5}&74.9\textcolor{gray}{$\pm$0.8}&75.7\textcolor{gray}{$\pm$1.0}&60.0\textcolor{gray}{$\pm$1.1}&6.3\\
        & GIN~\cite{GIN} & 65.8\textcolor{gray}{$\pm$4.5}&70.1\textcolor{gray}{$\pm$5.4}&58.0\textcolor{gray}{$\pm$4.4}&74.0\textcolor{gray}{$\pm$0.8}&75.3\textcolor{gray}{$\pm$1.6}&57.3\textcolor{gray}{$\pm$1.9}&8.8\\
        & GAT~\cite{GAT} & 66.2\textcolor{gray}{$\pm$2.6}&69.7\textcolor{gray}{$\pm$6.4}&58.5\textcolor{gray}{$\pm$3.6}&\bf{75.4\textcolor{gray}{$\pm$0.5}}&72.9\textcolor{gray}{$\pm$1.4}&\bf{60.9\textcolor{gray}{$\pm$1.8}}&6.8\\
        & GraphSAGE~\cite{GraphSAGE} & 69.6\textcolor{gray}{$\pm$1.9}&72.5\textcolor{gray}{$\pm$1.9}&59.2\textcolor{gray}{$\pm$4.4}&74.7\textcolor{gray}{$\pm$0.7}&74.4\textcolor{gray}{$\pm$0.7}&60.4\textcolor{gray}{$\pm$1.0}&5.8\\
        \hline
        \multirow{4}*{Pre-train} & MolCLR~\cite{MolCLR} & 70.6\textcolor{gray}{$\pm$0.8}&75.5\textcolor{gray}{$\pm$0.9}&85.8\textcolor{gray}{$\pm$1.3}&70.7\textcolor{gray}{$\pm$1.8 }&72.7\textcolor{gray}{$\pm$2.1}&57.3\textcolor{gray}{$\pm$0.8}&6.3\\
        &Infomax~\cite{DGI} & 68.8\textcolor{gray}{$\pm$0.8}&75.9\textcolor{gray}{$\pm$1.6}&69.9\textcolor{gray}{$\pm$3.0}&75.3\textcolor{gray}{$\pm$0.5}&76.0\textcolor{gray}{$\pm$0.8}&58.4\textcolor{gray}{$\pm$0.7}&4.5\\
         &GraphCL~\cite{GraphCL} & 61.9\textcolor{gray}{$\pm$0.32}&72.0\textcolor{gray}{$\pm$0.8}&82.7\textcolor{gray}{$\pm$0.5}&69.9\textcolor{gray}{$\pm$0.8}&75.3\textcolor{gray}{$\pm$0.6}&57.0\textcolor{gray}{$\pm$0.7}&8.7\\
        &AttrMasking~\cite{Hu} & 67.5\textcolor{gray}{$\pm$0.9}&78.0\textcolor{gray}{$\pm$1.7}&73.4\textcolor{gray}{$\pm$1.3}&73.5\textcolor{gray}{$\pm$0.7}&76.9\textcolor{gray}{$\pm$0.8}&57.9\textcolor{gray}{$\pm$1.1}&5.3\\

        \hline
            
        \multirow{3}*{Prompt}
        &MolCLR+MolCPT&\bf{70.9\textcolor{gray}{$\pm$1.9}}&80.0\textcolor{gray}{$\pm$0.6}&\bf{87.1\textcolor{gray}{$\pm$0.9}}&71.3\textcolor{gray}{$\pm$0.7}&75.8\textcolor{gray}{$\pm$1.5}&60.0\textcolor{gray}{$\pm$1.0}&\bf{3.3}\\
        & GraphCL+MolCPT&67.7\textcolor{gray}{$\pm$0.7}&75.0\textcolor{gray}{$\pm$0.7}&78.2\textcolor{gray}{$\pm$0.6}&72.7\textcolor{gray}{$\pm$0.5}&75.8\textcolor{gray}{$\pm$0.9}&59.5\textcolor{gray}{$\pm$0.3}&5.8\\
       
        &  AttrMasking+MolCPT&68.8\textcolor{gray}{$\pm$0.5}&\bf{80.8\textcolor{gray}{$\pm$1.0}}&80.3\textcolor{gray}{$\pm$1.4}&74.3\textcolor{gray}{$\pm$0.5}&\bf{77.5\textcolor{gray}{$\pm$0.7}}&57.0\textcolor{gray}{$\pm$0.4}&4.2\\

        \bottomrule
        \end{tabular}
        \label{tab:main_results}
\end{table*}

\begin{table*}[t]
    \setlength{\tabcolsep}{9pt}
        \centering
        \caption{Ablation studies on the motif prompting and answer searching modules of MolCPT. AS stands for Answer Searching only, MP stands for Motif Prompting only.} 
        \begin{tabular}{ll|cccccccc}
        \toprule
        Prompt & \multirow{2}*{Models} & \multicolumn{6}{c}{Datasets} & \\
         Methods&  & \textbf{BBBP} & \textbf{BACE} & \textbf{ClinTox} & \textbf{Tox21} & \textbf{HIV} & \textbf{SIDER} & Avg. ranking\\
        \midrule
        \multirow{3}*{AS Only} 
        &MolCLR & 71.9\textcolor{gray}{$\pm$2.2}&80.4\textcolor{gray}{$\pm$2.7}&76.4\textcolor{gray}{$\pm$1.4}&71.1\textcolor{gray}{$\pm$1.0}&74.8\textcolor{gray}{$\pm$0.9}&59.3\textcolor{gray}{$\pm$1.0}&5.7\\
        & GraphCL &69.6\textcolor{gray}{$\pm$0.8}&74.9\textcolor{gray}{$\pm$0.6}&73.0\textcolor{gray}{$\pm$0.9}&71.7\textcolor{gray}{$\pm$0.8}&75.3\textcolor{gray}{$\pm$1.0}&\bf{61.0\textcolor{gray}{$\pm$0.7}}&5.8\\
        & AttrMasking & 68.7\textcolor{gray}{$\pm$1.1}&80.8\textcolor{gray}{$\pm$0.9}&73.9\textcolor{gray}{$\pm$1.0}&74.2\textcolor{gray}{$\pm$0.5}&\bf{77.6\textcolor{gray}{$\pm$1.1}}&57.9\textcolor{gray}{$\pm$0.9}&4.7\\

        \hline
        \multirow{3}*{MP Only}
        &MolCLR & \bf{72.6\textcolor{gray}{$\pm$1.6}}&77.0\textcolor{gray}{$\pm$1.5}&\bf{89.3\textcolor{gray}{$\pm$1.1}}&71.3\textcolor{gray}{$\pm$0.9}&73.3\textcolor{gray}{$\pm$1.0}&60.0\textcolor{gray}{$\pm$0.6}&4.7\\
        &GraphCL & 67.2\textcolor{gray}{$\pm$0.7}&70.8\textcolor{gray}{$\pm$0.5}&80.3\textcolor{gray}{$\pm$0.7}&73.2\textcolor{gray}{$\pm$0.4}&76.4\textcolor{gray}{$\pm$0.8}&58.9\textcolor{gray}{$\pm$0.7}&6.2\\
        & AttrMasking & 67.4\textcolor{gray}{$\pm$0.7}&77.9\textcolor{gray}{$\pm$0.6}&82.8\textcolor{gray}{$\pm$0.7}&\bf{74.3\textcolor{gray}{$\pm$0.4 }}&76.6\textcolor{gray}{$\pm$0.1}&58.3\textcolor{gray}{$\pm$0.3}&4.5\\

        \hline
            
        \multirow{3}*{MolCPT}&MolCLR&70.9\textcolor{gray}{$\pm$1.3}&80.0\textcolor{gray}{$\pm$1.9}&87.1\textcolor{gray}{$\pm$0.6}&71.3\textcolor{gray}{$\pm$0.9}&75.8\textcolor{gray}{$\pm$0.7}&60.0\textcolor{gray}{$\pm$1.0}&\bf{3.8}\\
        & GraphCL&67.7\textcolor{gray}{$\pm$0.7}&75.0\textcolor{gray}{$\pm$0.7}&78.2\textcolor{gray}{$\pm$0.6}&72.7\textcolor{gray}{$\pm$0.5}&75.8\textcolor{gray}{$\pm$0.9}&59.5\textcolor{gray}{$\pm$0.3}&5.8\\
       
        &  AttrMasking&68.8\textcolor{gray}{$\pm$0.5}&\bf{80.8\textcolor{gray}{$\pm$1.0}}&80.3\textcolor{gray}{$\pm$1.4}&74.3\textcolor{gray}{$\pm$0.5}&77.5\textcolor{gray}{$\pm$0.7}&57.0\textcolor{gray}{$\pm$0.4}&\bf{3.8}\\
        
        \bottomrule
        \end{tabular}
        \label{tab:ablation_modules}

\end{table*}

\section{Experiments}
\label{sec:experiments}
We investigate how MolCPT improves pre-trained GNN performance on molecular property prediction tasks. In our investigations, we raise the following questions: \textbf{Q1:} Compared with supervised and pre-trained baselines, how effective is MolCPT in transfer learning settings? \textbf{Q2:} How much of MolCPT's gains can be attributed to answer searching or motif prompting? \textbf{Q3:} Can MolCPT prompt task-relevant information out of frozen pre-trained GNNs? \textbf{Q4:} How well does MolCPT, and answer searching in general, perform in zero-shot prediction settings? \textbf{Q5:} How does the choice of motif vocabulary affect downstream performance, with and without the vocabulary filtering step? 

\subsection{Datasets} We evaluate MolCPT on six molecular classification datasets contained in MoleculeNet~\cite{MoleculeNet}, namely BBBP, BACE, ClinTox, Tox21, HIV, and SIDER. More details for these datasets are described in Appendix \ref{app:summary_stats}.

\subsection{Baselines}
To evaluate MolCPT, we compare with two categories of baselines: supervised approaches without pre-training, and pre-trained approaches without prompting. 

\noindent\textbf{Supervised baselines.} We evaluate state-of-the-art GNNs widely used for molecular property prediction, and train them only using the labeled dataset. Specifically, we evaluate GCN~\cite{GCN}, GIN~\cite{GIN}, GAT~\cite{GAT}, and GraphSAGE~\cite{GraphSAGE}. 
    
\noindent\textbf{Pre-trained baselines.} We evaluate four popular pre-training approaches, including (i) Infomax~\cite{DGI} maximizing the mutual information between subgraphs and their corresponding graphs, (ii) MolCLR~\cite{MolCLR} leveraging node, edge, and motif elements to generate graph views for contrastive learning, (iii) GraphCL~\cite{GraphCL} contrasting graphs and their randomly perturbed counterparts, and (ii) AttrMasking~\cite{Hu} predicting masked node attributes. Each pre-training baseline is fine-tuned using the standard procedure of linear probing, in which a linear layer is applied after the pre-trained GNN to make categorical predictions.

\subsection{Implementation of MolCPT}
To demonstrate the flexibility of MolCPT, we choose three pre-trained baselines as backbone models for our plug-in prompt technique: AttrMasking, GraphCL, and MolCLR. For fair comparison, we evaluate all pre-trained baselines using the experiment settings provided by \citet{GraphCL}. The exceptions are AttrMasking, which specifies a dropout rate of 0.5, and GraphCL, for which we only evaluate the pre-trained GNN trained for 100 epochs. Otherwise, we use the same pre-training and fine-tuning hyperparameters, as well as the 5-layer GIN backbone. For motif prompting, we tune the key hyperparameters of motif filtering threshold $t$ and number of heads. For answer searching, we tune the answer ensemble size and orthogonal weight penalty, described by \citet{GPPT}. More details on our hyperparameter search procedure are given in Appendix \ref{hp_search}.

\subsection{Molecule Property Prediction Studies}
We compare MolCPT with the baseline approaches in Table \ref{tab:main_results} to answer \textbf{Q1}. We make the following observations:

\textbf{\ding{182}} \textit{MolCPT significantly boosts model performances for molecular property prediction}. First, the pre-trained approaches generally outperform the supervised ones by encoding prior topological knowledge. MolCPT further generalizes the pre-trained models for more effective transfer learning. In particular, compared with MolCLR, MolCPT improves average test scores by \textbf{2.1\%} across the 6 tasks while beating MolCLR on 4 of them. Compared with vanilla GraphCL, MolCPT improves average test scores by \textbf{1.7\%} and beats GraphCL on 5 of them. Compared with vanilla AttrMasking, MolCPT improves average test scores by \textbf{1.9\%} and beats AttrMasking on 5 of them. These results empirically validate the MolCPT's gains are agnostic to the choice of pre-training approach. One can easily reuse the same pre-training objective, and quickly extend it to the downstream problem by deploying our prompting scheme.

\textbf{\ding{183}} \textit{MolCPT reduces negative transfer resulting from the misaligned pre-training and fine-tuning objectives}. Similar to \citet{Sun}, we observe significant negative transfer from the pre-trained baselines, across the 6 datasets. \citet{GraphCL} attribute this problem to ill-posed graph augmentations that corrupt chemical structure during pre-training. \citet{Sun} observe more generally that pre-training with random topological masking may not align with downstream molecular analysis. MolCPT bridges the misalignment between pre-training and fine-tuning using motif prompting and answer searching; table \ref{tab:main_results} demonstrates that MolCPT resolved 5 out of 7 instances of negative transfer from the pre-trained models. This supports our hypothesis that well-designed prompts can learn semantic structure information for downstream applications.

\begin{figure}[H]
    \centering
    \includegraphics[width=0.9\linewidth]{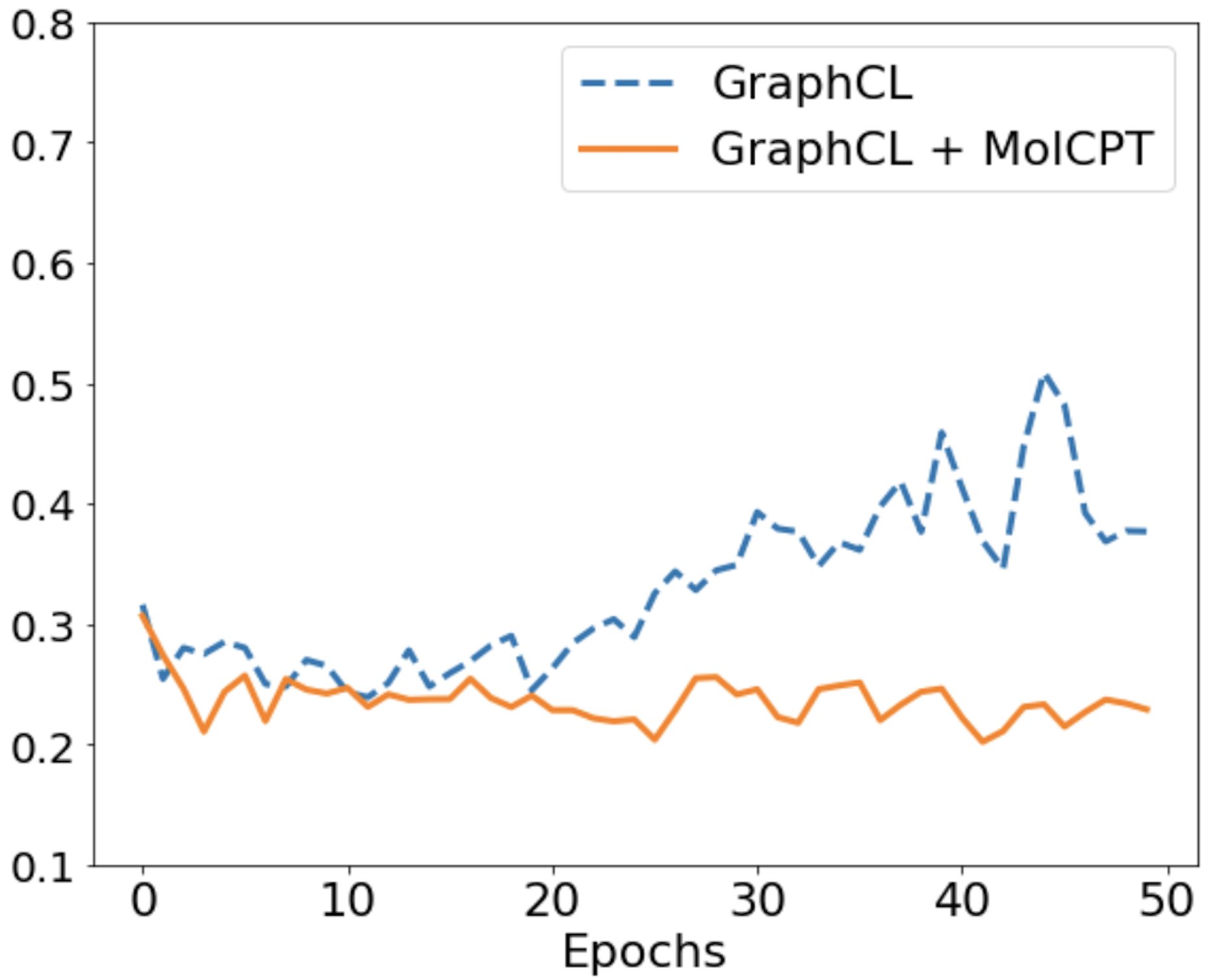}
    \caption{Validation losses of GraphCL and GraphCL + MolCPT on the Tox21 dataset.
    }
    \label{fig:training_dynamics}
\end{figure}

\textbf{\ding{184}} \textit{MolCPT stabilizes convergences of pre-trained baselines}. We observed that the GraphCL baseline is especially prone to unstable convergences; it reaches its minimum validation loss early on before its performance rapidly degrades. With MolCPT, GraphCL performance remains stable at the minimum validation loss throughout training. We illustrate this phenomenon more clearly in Figure \ref{fig:training_dynamics}.

\begin{table*}[t]
    \setlength{\tabcolsep}{10pt}
        \centering
        \caption{ Results comparing models with frozen pre-trained GNN, measured in ROC-AUC (\%). MolCPT beats baselines on all 6 classification tasks.}  
        \begin{tabular}{ll|cccccccc}
        \toprule
        \multirow{2}*{Frameworks} & \multirow{2}*{Models} & \multicolumn{6}{c}{Datasets} & \\
        &  & \textbf{BBBP} & \textbf{BACE} & \textbf{ClinTox} & \textbf{Tox21} & \textbf{HIV} & \textbf{SIDER} \\
        \midrule
            \multirow{2}*{Pre-train}&MolCLR~\cite{MolCLR}&62.9\textcolor{gray}{$\pm$0.5}&61.0\textcolor{gray}{$\pm$0.4}&55.3\textcolor{gray}{$\pm$0.5 }&68.0\textcolor{gray}{$\pm$0.3}&64.5\textcolor{gray}{$\pm$0.2}&51.2\textcolor{gray}{$\pm$0.4}\\
            &GraphCL~\cite{GraphCL}&59.9\textcolor{gray}{$\pm$0.1}&70.6\textcolor{gray}{$\pm$0.0}&68.1\textcolor{gray}{$\pm$0.1}&70.6\textcolor{gray}{$\pm$0.1}&70.5\textcolor{gray}{$\pm$0.1}&54.6\textcolor{gray}{$\pm$0.0}\\
        \midrule
            \multirow{2}*{Prompt}&MolCLR + MolCPT&62.2\textcolor{gray}{$\pm$0.5}&72.3\textcolor{gray}{$\pm$0.7}&\bf{71.8\textcolor{gray}{$\pm$0.9}}&66.8\textcolor{gray}{$\pm$0.5}&63.2\textcolor{gray}{$\pm$1.0}&56.2\textcolor{gray}{$\pm$0.3}\\
            &GraphCL + MolCPT&\bf{63.7\textcolor{gray}{$\pm$0.5}}&\bf{75.3\textcolor{gray}{$\pm$0.8}}&65.3\textcolor{gray}{$\pm$0.4}&\bf{70.6\textcolor{gray}{$\pm$0.4}}&\bf{76.5\textcolor{gray}{$\pm$0.7}}&\bf{58.8\textcolor{gray}{$\pm$0.3}}\\

        \bottomrule
        \end{tabular}
        \label{tab:frozen_gnn}

\end{table*}

\begin{table*}[t]
    \setlength{\tabcolsep}{7pt}
        \centering
        \caption{ Results comparing models in the zero-shot setting, measured in ROC-AUC (\%). AS means Answer Search. }.  
        \begin{tabular}{ll|ccccccccc}
        \toprule
        \multirow{2}*{Frameworks} & \multirow{2}*{Models} & \multicolumn{6}{c}{Datasets} & \\
        &  & \textbf{BBBP} & \textbf{BACE} & \textbf{ClinTox} & \textbf{Tox21} & \textbf{HIV} & \textbf{SIDER} & Avg. ranking\\
        \midrule
            \multirow{2}*{Pre-train}&MolCLR~\cite{MolCLR}&51.2\textcolor{gray}{$\pm$1.3}&44.6\textcolor{gray}{$\pm$4.3}&49.9\textcolor{gray}{$\pm$2.5 }&48.0\textcolor{gray}{$\pm$2.7}&54.1\textcolor{gray}{$\pm$3.6}&49.8\textcolor{gray}{$\pm$4.7}&5\\
            &GraphCL~\cite{GraphCL}&45.2\textcolor{gray}{$\pm$0.0}&40.2\textcolor{gray}{$\pm$0.0}&43.5\textcolor{gray}{$\pm$0.0}&49.3\textcolor{gray}{$\pm$0.0}&62.6\textcolor{gray}{$\pm$0.0}&50.9\textcolor{gray}{$\pm$0.0}&4.5\\
        \midrule
            \multirow{4}*{Prompt}&MolCLR + AS Only&57.0\textcolor{gray}{$\pm$2.0}&64.7\textcolor{gray}{$\pm$1.4}&\bf{55.6\textcolor{gray}{$\pm$1.7}}&61.3\textcolor{gray}{$\pm$2.6}&60.7\textcolor{gray}{$\pm$2.5}&45.4\textcolor{gray}{$\pm$1.7}&3\\
            &GraphCL + AS Only&\bf{62.5\textcolor{gray}{$\pm$1.5}}&62.2\textcolor{gray}{$\pm$1.1}&48.4\textcolor{gray}{$\pm$0.9}&64.6\textcolor{gray}{$\pm$2.1}&63.3\textcolor{gray}{$\pm$0.3}&\bf{54.0\textcolor{gray}{$\pm$0.8}}&\bf{2.2}\\
            &MolCLR + MolCPT&56.4\textcolor{gray}{$\pm$3.2}&59.7\textcolor{gray}{$\pm$2.3}&53.2\textcolor{gray}{$\pm$2.9}&49.4\textcolor{gray}{$\pm$2.1}&53.4\textcolor{gray}{$\pm$2.2}&48.7\textcolor{gray}{$\pm$1.2}&4\\
            &GraphCL + MolCPT&58.7\textcolor{gray}{$\pm$2.1}&\bf{69.1\textcolor{gray}{$\pm$2.7}}&50.6\textcolor{gray}{$\pm$2.1}&\bf{64.6\textcolor{gray}{$\pm$0.9}}&\bf{65.2\textcolor{gray}{$\pm$1.4}}&43.0\textcolor{gray}{$\pm$0.9}&2.3\\

        \bottomrule
        \end{tabular}
        \label{tab:zero_shot}

\end{table*}

\subsection{Motif Prompting and Answer Searching}
\label{module_ablations}
To answer \textbf{Q2}, we evaluate the two modules that comprise MolCPT: motif prompting and answer searching. We observe:

\textbf{\ding{185}} \textit{MolCPT outperforms motif prompting and answer searching in isolation}. Motif prompting and answer searching serve independent purposes: motif prompting augments pre-trained molecular representations, and answer searching bridges pre-training and fine-tuning objectives. Both modules must be employed for effective transfer learning, and no one module is solely responsible for MolCPT's impressive gains. We point out that motif prompting and answer searching, in isolation, still improve upon standard fine-tuning performance. Answer searching achieves more significant gains in the zero-shot setting (Section \ref{zero_shot}).

\subsection{Freezing Weights}
We investigate \textbf{Q3} by freezing the pre-trained weights, to evaluate whether MolCPT can adapt the frozen model to diverse downstream tasks. This mimics real-world scenarios, in which one may prefer not to update the parameters involved in complex graph convolutions. We report the results of our experiments in Table~\ref{tab:frozen_gnn} and make the following observation:

\textbf{\ding{186}} \textit{MolCPT enables accurate prediction of molecular properties, without needing to finetune the backbone GNN or projection head}.
With frozen pre-trained weights, MolCPT improves average test scores by \textbf{5.1\%} above MolCLR and \textbf{2.6\%} above GraphCL. These results agree with conventional wisdom that prompting successfully ``prompts'' task-relevant information out of the pre-trained model. 

\subsection{Zero-Shot Prediction}
\label{zero_shot}
To answer \textbf{Q4}, we evaluate the baseline approaches, MolCPT, and our general answer searching scheme (without motif prompting) in the zero-shot setting. Our results are reported in Table \ref{tab:zero_shot}, from which we observe:

\textbf{\ding{187}} \textit{Our answer searching scheme turns the pre-trained model into an effective zero-shot predictor}. First, we point out that the standard fine-tuning procedure makes close-to-random predictions in the zero-shot setting. This is because the fine-tuning weights must be trained to learn the downstream task. Answer searching, on the other hand, designs an answer space the pre-trained model can easily infer over \textit{with no training}. Thus, on MolCLR and GraphCL baselines, our answer searching scheme beats average test scores of baselines \textbf{7.9\%} and \textbf{10.6\%}, respectively. MolCPT slightly underperforms the scheme because the motif prompting module requires training to properly weight motif embeddings. However, MolCPT still beats MolCLR and GraphCL average test scores by \textbf{3.9\%} and \textbf{9.9\%}, respectively.

\begin{figure}
    \centering
    \includegraphics[width=\linewidth]{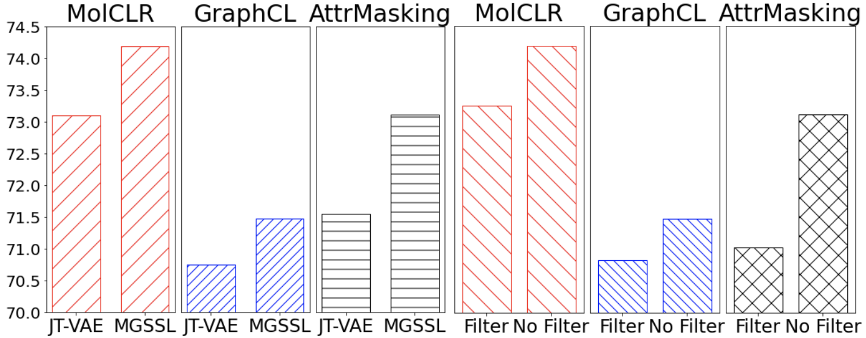}
    \caption{Ablation studies of motif detection method, averaging over all datasets.
    }
    \label{fig:vocab_size}
\end{figure}

\begin{figure}
    \centering
    \includegraphics[width=0.9\linewidth]{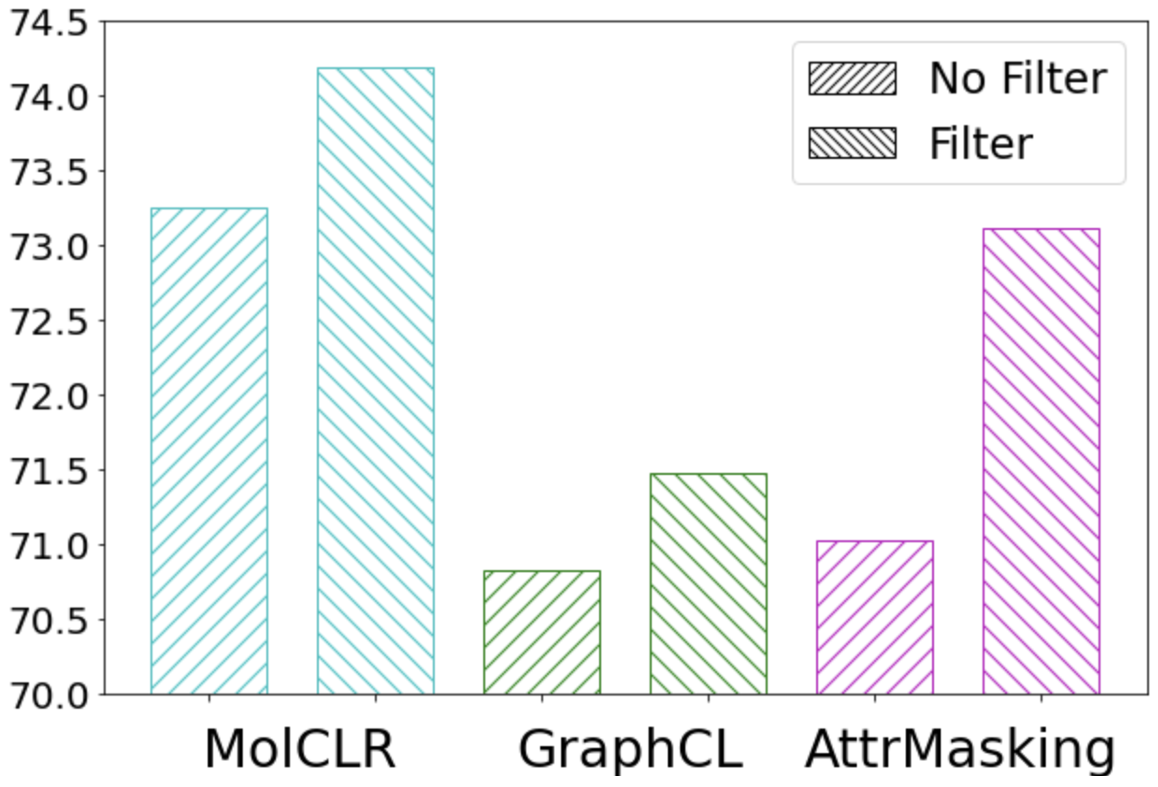}
    \caption{Ablation studies of motif filtering threshold, averaging over all datasets.
    }
    \label{fig:filter_size}
\end{figure}

\subsection{Number of Key Motifs}
To answer \textbf{Q5}, we study two key components affecting the size of motif vocabulary in MolCPT: the graph fragmentation method, and the filtering threshold. By ablating these components on GraphCL and MolCLR, we make the following observations (visualized in Figure \ref{fig:vocab_size}):


\textbf{\ding{188}} \textit{The graph fragmentation method heavily affects downstream task performance}. We evaluate two different fragmentation methods, in order of increasing granularity: JT-VAE and MGSSL. Granularity is an important consideration when generating the motif vocabulary: MGSSL finds that coarse-grained motifs have lower occurrence frequencies, preventing the model from learning motif embeddings suitable for downstream application. We find that the finer granularity of MGSSL produces the most informative motifs.

\textbf{\ding{189}} \textit{The filtering threshold improves MolCPT's generalizations to downstream tasks}. Without filtering, the motif vocabulary contains rare motifs whose embeddings are updated on very few molecules. These rare motif embeddings cannot learn semantic structures, and thus fail to properly indicate molecular properties. 

\begin{figure}
    \centering
    \includegraphics[width=0.9\linewidth]{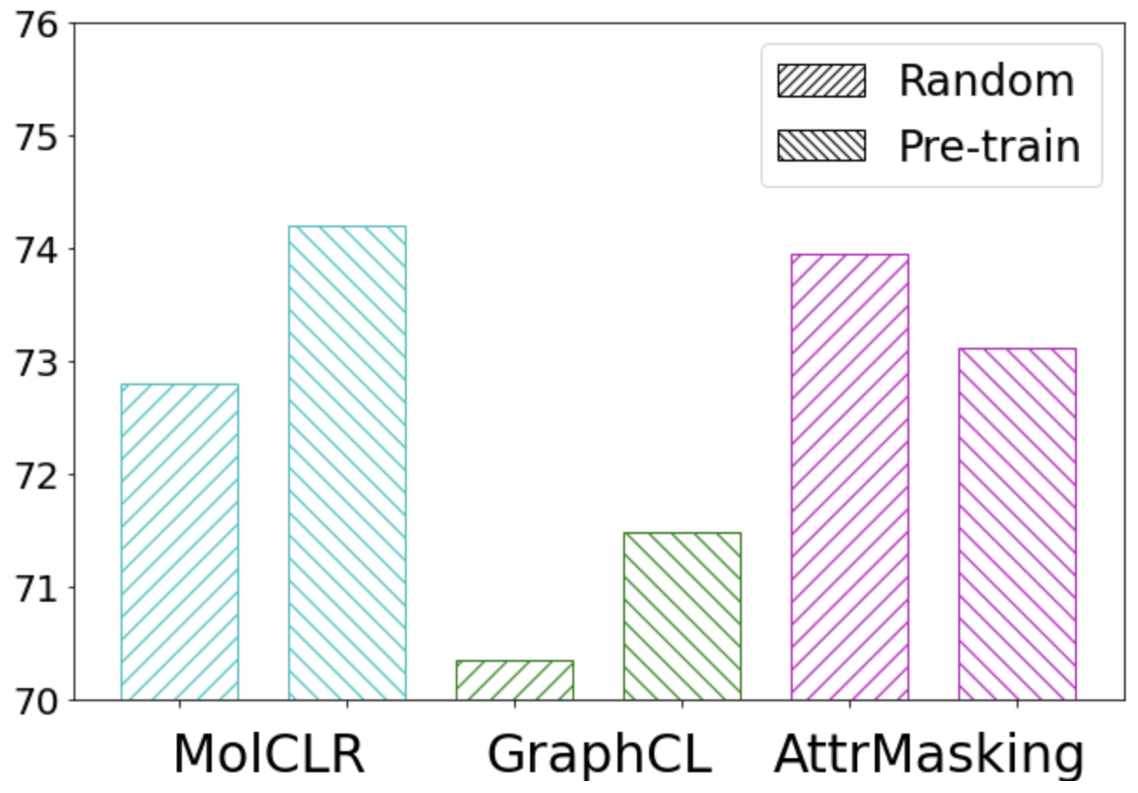}
    \caption{Ablation studies of prompt initialization strategy, averaging over all datasets.
    }
    \label{fig:init_strat}
\end{figure}

\subsection{Prompt Initialization Method}
We investigate \textbf{Q6} by comparing our initialization strategy (using pre-trained motif embeddings) to random initialization. For random initialization, we use uniform Xavier initialization \citep{XavierInit} on all motif embeddings except the empty motif. We show our ablation results in Figure \ref{fig:init_strat}, and observe:

\textbf{\ding{190}} \textit{Choosing the right initialization method is crucial for learning informative motif embeddings}. As stated in Section \ref{embedding_table}, random initialization incorrectly distinguishes between different motifs at the starting phase. Thus, motif prompting fails to learn semantic knowledge required for identifying molecular properties. Our strategy of initializing the motif embeddings by inferring their subgraph embeddings is simple, but surprisingly effective.

\section{Conclusion}
We propose MolCPT, the first ``pre-train, prompt, fine-tune" method for molecular property prediction. Specifically, MolCPT uses the motif prompting function and differentiable answer search to bridge the gap between pre-training and fine-tuning objectives. We design the prompting function to augment the input molecule using key motifs, in a way that captures the motif structure without corrupting the molecular graph. MolCPT also adopts a proper initialization method, attention network, and constrained learning to further encode molecular representations with task-relevant structure information. Given the diverse pre-training tasks used in molecular property prediction, we design a differentiable answer search strategy to map the latent pre-training outputs to desired property labels. By reformulating the downstream applications look like pre-training one, the molecular property prediction is free to use any of the trained GNNs. Extensive experiments on six benchmark datasets show the generality, effectiveness, and efficiency of MolCPT for improving performances over the existing pre-trained molecular models.

\bibliographystyle{ACM-Reference-Format}
\bibliography{sample-base}

\newpage

\appendix

\section{Dataset Statistics}
\label{app:summary_stats}

The summary statistics for MoleculeNet datasets are shown in Table \ref{tab:data_summary_stats}. We do train/validation/test splitting of datasets using the scaffold-split procedure, recommended by~\cite{MoleculeNet}. 

\begin{table}[h]
\renewcommand{\arraystretch}{1.0}
\caption{Summary Statistics of MoleculeNet Datasets
}
\label{tab:data_summary_stats}
\vspace{-10pt}
\begin{center}
\begin{tabular}{ l | r | r }\hline
{\textbf{Dataset}} & {\textbf{\# Tasks}} & {\textbf{\# Compounds}}  \\ \hline

BBBP & 1 & 2053 \\

BACE & 1 & 1522 \\

ClinTox & 2 & 1491 \\

Tox21 & 12 & 8014 \\

HIV & 1 & 41913 \\

SIDER & 27 & 1427 \\

\hline

\end{tabular}
\end{center}
\end{table}

\section{Hyperparameter Search Procedure}
\label{hp_search}

To tune the hyperparameters of MolCPT, we used Bayesian Optimization as implemented by the hyperopt package \citep{hyperopt}. Searches are terminated after 50 iterations. 

\begin{itemize}
    \item For motif prompting threshold $t$, we searched along $[0, 100]$ with step size $10$
    \item For number of heads, we searched along $[2, 4, 8]$ for MolCLR and $[2, 4, 10]$ for GraphCL and AttrMasking
    \item For answer ensemble size, we searched along $[0, 100]$ with step size $2$
    \item For orthogonal weight penalty, we searched along $[0, 10^{-4}]$ with step size $5 \cdot 10^{-6}$
\end{itemize}

For fair comparison with ablations (Section \ref{module_ablations}), we first tuned hyperparameters of the motif prompting module, before tuning hyperparameters of the answer searching module.

\end{document}